\documentclass[10pt, a4paper]{article}

\usepackage{lrec2026} 

\usepackage{subfigure}
\usepackage{amsmath}
\usepackage{makecell}
\usepackage{booktabs}
\usepackage{multirow}
\usepackage{array}

\newcommand{\ignore}[1]{}

\title{Lexicalized Constituency Parsing for Middle Dutch: Low-resource Training and Cross-Domain Generalization}

\name{Yiming Liang, Fang Zhao} 

\address{Universiteit Gent, Université Paris Cité \& Laboratoire de linguistique formelle \\
         yiming.liang@ugent.be, fang.zhao@etu.u-paris.fr }

\abstract{
Recent years have seen growing interest in applying neural networks and contextualized word embeddings to the parsing of historical languages. However, most advances have focused on dependency parsing, while constituency parsing for low-resource historical languages like Middle Dutch has received little attention. In this paper, we adapt a transformer-based constituency parser to Middle Dutch, a highly heterogeneous and low-resource language, and investigate methods to improve both its in-domain and cross-domain performance. We show that joint training with higher-resource auxiliary languages increases F1 scores by up to 0.73, with the greatest gains achieved from languages that are geographically and temporally closer to Middle Dutch. We further evaluate strategies for leveraging newly annotated data from additional domains, finding that fine-tuning and data combination yield comparable improvements, and our neural parser consistently outperforms the currently used PCFG-based parser for Middle Dutch. We further explore feature-separation techniques for domain adaptation and demonstrate that a minimum threshold of approximately 200 examples per domain is needed to effectively enhance cross-domain performance. 
 \\ \newline }

\begin{document}

\maketitleabstract

\section{Introduction}

Neural networks with contextualized word representations have proven effective for PoS tagging and syntactic parsing \cite[][among others]{kitaev2018Constituency,han2019Unsupervised,lim2020SemiSupervised}, reducing the manual effort needed to build linguistic treebanks, which are crucial for quantitative linguistic studies and digital humanities research.
Recently, there has been a growing interest in applying them to parse lesser-resourced languages like historical languages (\citealp[e.g.,][for Old French]{grobol2021Analyse}; \citealp[][for Early Modern English]{kulick2022PennHelsinki}).

While most advances have focused on dependency parsing \cite[e.g.,][]{vania2019Systematic,zhang2022Neural,grobol2022BERTrade}, much less attention has been paid to constituency parsing, as the latter one has been argued to be more difficult in terms of accuracy and the efficiency of parsing algorithms \cite{kubler2009Dependencya,cross2016SpanBased}. 
However, constituency-parsed corpora remain widely used in linguistic research, in particular among historical linguists, because they provide the hierarchical structure of sentences in terms of smaller units that better align with many influential syntactic frameworks (e.g., traditional Phrase-Structure Grammar \cite{bloomfield1933Language}, Generative Grammar \cite{chomsky1957Syntactica}), and are easily accessed and explored with the toolkit \textit{CorpusSearch 2} \footnote{https://corpussearch.sourceforge.net/CS.html} \cite{randall2010CorpusSearch} provided by the University of Pennsylvania.
Constituency parsing has also been shown to provide useful structured input for downstream NLP tasks and improve performance in lots of semantic tasks, such as Semantic Role Labeling \cite{wang2019How,bastianelli2020Encoding}, Target Identification \cite{bastianelli2020Encoding} and nested Named Entity Recognition \cite{wang2018Neural,fu2021Nested,yang2022BottomUp}.

Despite its importance, there has been little work on applying transformer-based models and contextualized word representations to historical language constituency parsing \cite[but see][]{kulick2022PennHelsinki,sapp2023Parsing,nie2023CrossLinguala}, and no work has been focused on Middle Dutch parsing. The majority of Penn-style historical corpora still rely on rule-based shallow parsing or statistical PCFG parsers, which require extensive manual effort for postprocessing and correction \cite{booth2020Pennstyle,farasyn2022Challenges}.
In this paper, we focus on constituency parsing of Middle Dutch (1150-1500), a low-resource language with only around 4,000 syntactically annotated sentences of high heterogeneity in spelling, syntax and genres.
We apply the Berkeley Neural parser (\textit{Benepar}, \citealt{kitaev2018Constituency,kitaev2019Multilingual}) combined with BERT representations \cite{devlin2019BERT} to Middle Dutch, and investigate how annotated corpora from richer-resourced languages and domain-adaptation techniques can be leveraged to train a constituency parser that performs well both in-domain and on out-of-domain texts under low-resource conditions.
Concretely, first, we train \textit{Benepar} on the largest available parsed text  \textit{Etstoel} (approximately 2,000 sentences), and examine whether incorporating PoS-tag prediction as an auxiliary task \textbf{(Phase I)} and joint training with auxiliary languages can improve in-domain performance \textbf{(Phase II}). Second, based on the best-performing configuration, we explore its zero-shot and few-shot generalisation to new texts and genres, comparing data combination and fine-tuning strategies. We further experiment with feature separation techniques \cite{kim-etal-2016-frustratingly, li-etal-2020-semi, li-etal-2022-semi} to evaluate their role in low-resource constituency parsing (\textbf{Phase III}).
We also compare the performance of \textit{Benepar} with the statistical PCFG \textit{Bikel} parser \cite{bikel2002Design,bikel2004Intricacies}, currently in use for parsing Middle Dutch texts. 

We address two main research questions:

\textbf{RQ1:} \textit{How to effectively train a neural constituency parser when annotated data are limited?}
We find that while including PoS-tag prediction as an auxiliary task brings no additional benefit, joint training with auxiliary languages substantially improves accuracy (up to +0.73 F1 and at best 86.21 F1), especially when the auxiliary language is geographically and temporally close to Middle Dutch.

\textbf{RQ2:} \textit{How well does the parser generalize to new texts and genres, and how to improve cross-domain performance under high heterogeneity?}
Our experiments show that \textit{Benepar} strongly outperforms the statistical \textit{Bikel} parser even in zero-shot conditions. Fine-tuning and data-combined retraining with sentences from the new domain
yield comparable results, and the parser begins to show improvement with as few as 10 examples and surpasses 70 F1 after 100 examples.
We further examine feature separation techniques for domain adaptation and find that, with around 200 examples per domain, cross-domain performance exceeds 74.7 F1 across all three new domains, while smaller datasets yield little or no improvement.


Overall, our results not only produce a state-of-the-art Middle Dutch parser, but also yield broader insights into neural constituency parsing for low-resource and highly heterogeneous historical languages, particularly in the context of incremental treebank construction, where annotated data become available over time\footnote{The code for our experiments is available via: https://anonymous.4open.science/r/Goeiemiddutch}.

\section{Related work}

\subsection{Constituency parsing}

\ignore{
Plan
\begin{itemize}
    \item Previous attempts of parsing historical data, all with Berkeley Neural parser \cite{kitaev2018Constituency,kitaev2019Multilingual} (henceforth \textit{Benepar}):
    \begin{itemize}
        \item IPCHG: trained on CHLG and use in IPCHG, mix-training of CHLG and high german, 460 training sentences, but f1 score below 50 (probably because of the size of training for high german
        \item PPCEME: train benepar on PPCEME \cite{kulick2022PennHelsinki}
        \item The Icelandic Parsed Historical Corpus: benepar trained on IcePaHC (84.74 F1) and develop a neural pipeline to process Icelandic corpora \cite{arnardottir2020Neural}.
        \item delexcialized parsing: benepar, historical german, zero shot from modern German to middle German \cite{nie2023CrossLinguala}
    \end{itemize}
\end{itemize}
}

Constituency parsing, which seeks to represent the structure of sentences in terms of hierarchical phrases/constituents, has seen continuous progress from statistical probabilistic context-free grammar (PCFGs) \cite{charniak1997Statistical,collins1999HeadDriven,bikel2004Intricacies} to neural approaches \cite{cross2016SpanBased,choe2016Parsing,crabbe2015Multilingual,coavoux2017Multilingual}, and more recently, to self-attentive models with contextualised word embeddings \cite{kitaev2018Constituency,kitaev2019Multilingual,zhou2019HeadDriven}. While these advances have led to highly accurate parsers for modern languages with abundant annotated data such as the Penn Treebank (PTB), much less attention has been given to historical or low-resource languages, where annotated corpora are scarce and domain variation is high. 
In fact, Penn-style historical treebanks are often created using rule-based shallow parsers, which rely heavily on manual effort for creating rules and performing corrections on the output \cite{booth2020Pennstyle,farasyn2022Challenges,arnardottir2020Neural}. 
Among the very few historical treebanks that use an automatic parser, the Historical High German corpus (IPCHG, \citealp{sapp2024Introducing}) uses \textit{Benepar} \cite{kitaev2018Constituency,kitaev2019Multilingual} for automatic parsing: they first train \textit{Benepar} in richer-resourced Middle Low German (CHLG, \citealt{booth2020Pennstyle}) and apply it to the target Early New High German (ENHG), and subsequently retrain the parser on combined data consisting of CHLG and the manually corrected ENHG trees \cite{sapp2023Parsing}. 
Their experiments yield promising results (F1 65 at best), but the out-of-domain performance is far from satisfactory (below F1 50 in general), and the cross-domain performance remains unknown.
Other research also explores \textit{Benepar} on large historical Penn-style corpora: 
\citet{arnardottir2020Neural} report that \textit{Benepar} achieves an F1 of 84.74 when trained and evaluated on Icelandic Parsed Historical Corpus (73,000 matrix clauses, \citealp{ rognvaldsson-etal-2012-icelandic});
\citet{kulick2022PennHelsinki} achieve an F1 of 90.53 with 31 function tags trained on Parsed Corpus of Early Modern English (28,000 sentences, \citealt{kroch2004PennHelsinki}).
\citet{nie2023CrossLinguala} propose a delexicalized version of \textit{Benepar} and report an F1 of 64.72 in a zero-shot setup, transferring a model trained on Modern High German (TIGER, \citealt{brants2004TIGER}) to Middle High German. 
However, no attempts have yet been made to apply constituency parsing to Middle Dutch or Historical Dutch more generally.

\subsection{Domain adaptation}
Since Middle Dutch collectively refers to several dialects spoken over several centuries and there was no standardized variety \cite{huning2009Middle}, it exhibits substantial variation in spelling and syntax across texts.
This makes domain adaptation particularly crucial for parsing Middle Dutch, as models can suffer from dramatic performance drop when new texts and genres differ from training data \cite{ramponi-plank2020surveyneuralunsupervisedda, marzinotto2019robustsp, joshi2018extendingdomain}. 
Therefore, we review domain adaptation techniques, with a particular attention to low-resource settings.



A long-standing strand of domain adaptation approaches rests on explicitly separating domain-invariant features from domain-specific ones (e.g., \citealp{daume-iii-2007-frustratingly, kim-etal-2016-frustratingly, sato-etal-2017-adversarial, li-etal-2020-semi, li-etal-2022-semi}).
In particular, the so-called \textit{shared-private} model places a \textit{shared} encoder for domain-invariant features alongside one or more \textit{private} encoders for domain-specific signals \cite{kim-etal-2016-frustratingly}.
\citet{sato-etal-2017-adversarial} first applies adversarial training \cite{goodfellow2014gan, ganin2015udabackprop} to learn domain-invariant features in parsing.
They design a gating mechanism to mix representations from the shared and private encoders, then fed to the parser network.
However, while these methods consistently improve performance in most cases, they are harmful when target data are scarce.
Indeed, target domain encoders and gates may not be well optimized when lacking data, which in turn harms generalization.

\citet{li-etal-2020-semi, li-etal-2022-semi} publish a series of studies tackling the low-resource setting in domain adaptation in dependency parsing.
Following \citet{bousmalis2016domainseparationnetworks}, \citet{li-etal-2020-semi, li-etal-2022-semi} enforce orthogonality constraints to encourage the domain-specific features to be mutually exclusive with the shared features to reduce redundancies in the shared and private feature spaces.
To alleviate underfitting due to the lack of target domain labeled data, \citet{li-etal-2020-semi} introduce fused target-domain word representations, which combine source and target private representations as the final domain-specific representations when the input word is from the target domain.
From a similar angle, \citet{li-etal-2022-semi} apply a dynamic matching network \cite{jang2019learningtransfer} on the shared-private model to let the target encoder mimick well-trained source features, leveraging the in-depth relevance of domain-specific encoders and thus alleviating target domain underfitting.
While these methods have proven effective for small datasets, their lowest-resource domain studied consists of 1,645 examples.
It is therefore interesting to explore the effectiveness of these methods in even lower-resource conditions like Middle Dutch, and the amount of target data required.

\section{Data}

\ignore{
Plan
\begin{itemize}
    \item general information about the parsed texts: preprocessing, automatic annotations, manual check
    \item geographic information about the Middle Dutch corpus
    \item characteristics of Middle Dutch (with examples) (potential challenges for constituency parsing)
    \begin{itemize}
        \item high complexity deep embeddings in legal charters
        \item no spelling standardisation
        \item free word order ? (V2?)
    \end{itemize}
    \item Data preprocessing: data removal, and split remaining data of these texts for train, dev and test
\end{itemize}
}

\subsection{Parsed Data of Middle Dutch}
\textit{Middle Dutch }is the term used for the language varieties used between approximately 1150 and 1500 in the territory covered nowadays by the Netherlands and Flanders region of Belgium. Due to the lack of standard variety of Dutch and the dominance of Latin or French in writing, administration and nobility, Middle Dutch is a collection of dialects spoken over several centuries, and featured by ``a huge variation in the grammatical structure, the pronunciation and the spelling'' \cite[][p. 257]{huning2009Middle}, which implies big challenges for domain adaptation of the parser from one text to another. 
At the morpho-syntactic level, as stated in \citet{kerckvoorde1993Introduction,huning2009Middle}, Middle Dutch is most characterized by: 1) a rigid V2 order in main clause and a more flexible word order in subordinate clauses, whereas the finite verb is placed at the end of the subordinate clause in Modern Dutch; 2) more flexibility of word order within a nouns phrase (NP) because of modifier postpositioning; 
3) a much richer inflection system with a four-case-inflectional paradigm on nouns, adjectives, articles and numerals, which disappeared from the 17th century. 
As no word embeddings are available for Middle Dutch, the language’s flexible word order and rich inflectional morphology may pose additional challenges for parsing and word representation, particularly when adapting a pre-trained BERT model trained on later stages of Dutch.
An additional complexity comes from the nested embedded clauses and long sentences featured by legal charters, one of the most well-documented textual genres for Middle Dutch.

Our experiments are based on the following Middle Dutch texts: 
\begin{itemize}
  \setlength\itemsep{0em}
    \item \textit{Etstoel}: a sampling of legal charters composed in the 15th century
    \item \textit{CRM14}: a sampling of legal charters composed in the 14th century
    \item \textit{Tafel} ("Tafel van den Kersten Ghelove"): written by Dirck van Delft in 1404, religious texts
    \item \textit{Trappen} ("Seven Trappen"): written by Jan van Ruusbroec at 1359-1362, religious texts
\end{itemize}

\begin{figure}[!ht]
\begin{center}
\includegraphics[width=\columnwidth]{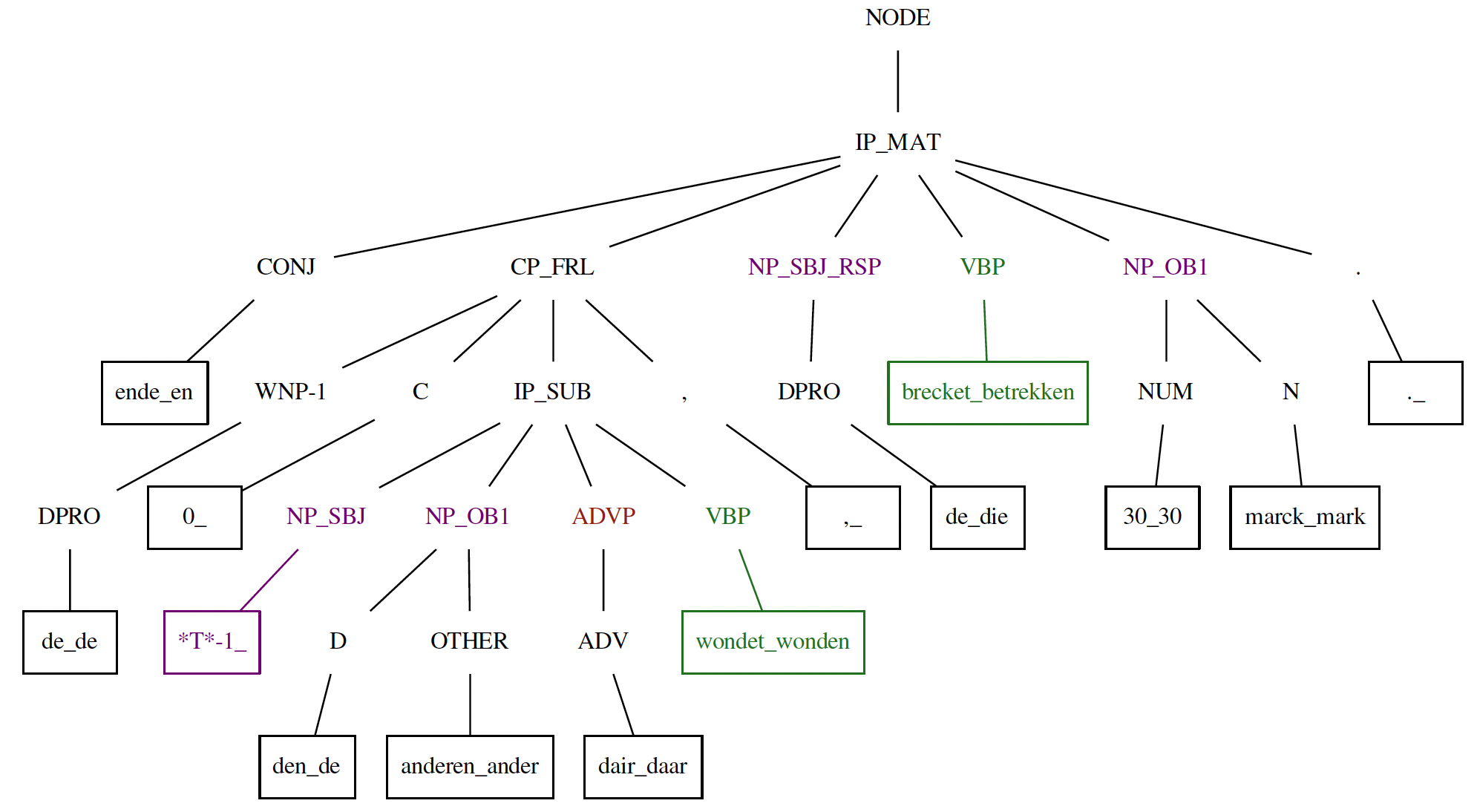}
\caption{A constituency tree from Etstoel.}
\label{fig:etstoel tree}
\end{center}
\end{figure}

A parsing tree from \textit{Etstoel }is illustrated by Figure \ref{fig:etstoel tree}. To the best of our knowledge, these are the only parsed texts of Middle Dutch.
More texts are being parsed and manually checked as part of a European Research Project. The four available texts are PoS-tagged and lemmatised either by being manually annotated by philologists, or using the GaLAHaD PoS-tagger and lemmatiser\footnote{https://github.com/instituutnederlandsetaal/galahad}, and parsed with Bikel parser\footnote{https://dbparser.github.io/dbparser/} \cite{bikel2002Design, bikel2004Intricacies}. 
All automatic annotations, including PoS tags, lemmas, and constituency trees, are subsequently corrected and carefully reviewed by human annotators. These texts are annotated following the Penn standard for historical English \cite{santorini2022Annotation}, with adaptations for Dutch-specific conventions.

\subsection{Data for Auxiliary Languages}
\label{sec:aux lang data}

It has been shown that cross-lingual transfer generally improves adaptation from high-resource modern languages to low-resource languages \cite{wolf-etal-2020-transformers, ruder2019neural, lauscher-etal-2020-zero, nie2023CrossLinguala, kitaev2019Multilingual}. 
Since Middle Dutch has limited annotated data, we experiment with the following Penn-style treebanks for auxiliary languages, differing in data sizes and relatedness to Middle Dutch: 
\begin{itemize}
    \item Historical treebanks: \\
        - Historical English (AllHE): Old English (OE, \citealt{taylor2003YorkTorontoHelsinki}), Middle English (ME, \citealt{kroch2000PennHelsinki}), Early Modern English (EModE, \citealt{kroch2004PennHelsinki}), Modern British English (ModBE, \citealt{kroch2016Penn})\\
        - Historical High German (AllHG): Middle High German (MHG), Early New High German (ENHG), both from \citet{sapp2024Introducing}\\
        Middle Low German (CHLG, \citealt{booth2020Pennstyle})\\
        - Historical French (AllHF): Old French (OF), Middle French (MF), both from \citet{france_martineau_modeliser_2010,kroch_penn_2010}\\
        - Historical Portuguese (HP, \citealt{galves2017Tycho})
    \item Modern treebanks: Modern English (PTB, \citealt{bies2015Penn}), Modern Chinese Treebank (CTB, \citealt{xue2013Chinese}), Modern Dutch (LassySmall, \citealt{vannoord2013Large})
\end{itemize}

All historical treebanks follow the Penn standard for Historical English \cite{santorini2022Annotation} for constituency parsing. Given the wide chronological span of historical languages and the syntactic changes that occur over time, the syntactic structures can differ substantially across historical periods. Therefore, most historical treebanks focus on one period. 
For corpora that are not explicitly divided, we separate them according to conventionally established periods, as in the cases of Historical High German and Historical French. 
Although modern treebanks also follow Penn-style, their syntactic tags and annotation conventions differ non-negligibly from those of historical treebanks. For this reason, we do not combine Modern British English and PTB, despite their temporal proximity.

\section{Experiments}

\subsection{Data Preparation}

For the first research question, which concerns low-resource parser training, we train the \textit{Benepar} parser only on \textit{Etstoel}, the largest parsed text. We explore two strategies: Part-of-Speech tags prediction as an auxiliary task (Phase I) and auxiliary language training (Phase II).
At the end of each phase, the best-performing option will be kept for the next phase.
After obtaining the best parser at the end of Phase II, we address the second research question, which addresses the generalization capacity of the parser to new texts and a new genre. For this purpose, we use the remaining three texts (CRM14, Tafel, and Trappen) in Phase III.

\subsubsection{Data preprocessing}

For all annotated data, including the auxiliary language corpora, empty categories and coreferential indices are removed. Sentences longer than 100 words are excluded, as they exceed the maximum sequence length supported by \textit{Benepar}. 
Unlike studies on constituency parsing in modern languages \citep[e.g.][]{cross2016SpanBased, kitaev2019Multilingual, coavoux2017Multilingual}, which remove all function tags during training and evaluation, we retain them because they provide crucial syntactic information essential for linguistic analysis.

\subsubsection{Evaluation metrics}
For \textbf{evaluation}, we use \textit{evalb}, a standard metric that compares spans and labels in gold and predicted trees, provided in the release of \citet{kitaev2019Multilingual}. Following \citet{kulick2022PennHelsinki} and \citet{sapp2023Parsing}, we report results with \textbf{all function tags} retained (e.g., “NP-SBJ” is treated as an atomic unit). To this end, we modified the \textit{evalb} source code to preserve function tags during evaluation. Punctuation is excluded from PoS tagging and parsing evaluation, following \citet{kitaev2018Constituency} and \citet{kitaev2019Multilingual}. 


\subsubsection{Cross-validation and resampling}
\label{sec:cross validation}
Since \textit{Etstoel} is relatively small, in Phases I and II, we apply 10-fold cross-validation to \textit{Etstoel} to enhance the reliability of our results .\footnote{See the suggestions of \citet{gorman-bedrick-2019-need}.}
For each run, eight folds are used for train, one fold serves as the dev set to determine the number of training epochs (early stopping), and one fold is reserved for test.
In Phase II, for each auxiliary language, the data are randomly split into 2,000 examples for dev, 2,000 examples for test, and the rest for train.
The same splits are used for all experiments.
As the main objective is to optimize the Middle Dutch parser, early stopping is used and the number of training epochs is determined based on the \textit{Etstoel} dev set defined in Phase I. All scores reported in Phases I and II for the parser are averaged across 10-fold cross-validation of \textit{Etstoel} test set.

\begin{table}[!ht]
\begin{center}
\footnotesize
\begin{tabular}{|l||c|c|c|c|}
     \hline
     \textbf{Dataset} & \textbf{Train} & \textbf{Dev} & \textbf{Test} & \textbf{Total} \\
     \hline
     \hline
     \textbf{Etstoel} & 1568 & 195 & 195 & 1958\\
     \textbf{CRM} & 10-100-200 & n/a & 169 & 369\\
     \textbf{Tafel} & 10-100-200 & n/a & 574 & 774\\
     \textbf{Trappen} & 10-100-200 & n/a & 751 & 951\\
     \hline
\end{tabular}
\caption{Middle Dutch data splits.\label{tab:splits}}
 \end{center}
\end{table}

In Phase III, for the zero-shot and fine-tuning experiments, the parser is retrained on nine folds of \textit{Etstoel} with the best auxiliary language to obtain the best in-domain model, while the remaining fold is used as the development set for early stopping.\footnote{The auxiliary language data is randomly resplit into 2,000 examples for development and the rest for training.}
For experiments that combine \textit{Etstoel} with other Middle Dutch data (combined data and adversarial training), the same nine folds of \textit{Etstoel} are used for training and the remaining fold for development.
As for the three remaining texts (considered as three new domains), since they are much smaller than \textit{Etstoel}, a 10-fold cross-validation is problematic due to limited test data per fold. Therefore, following \citet{sapp2023Parsing}, we use 10-resampling instead. In particular, for each text, 200 sentences are taken for training, and the remaining sentences are used for testing, with the sampling process repeated 10 times under each condition.
Because of the limited data size, no separate development set is used for the new texts; early stopping is again determined based on the \textit{Etstoel} dev set defined in Phase II.\footnote{Models not using \textit{Etstoel} (fine-tuning models in Phase III) are trained for an arbitrary of 50 epochs.} 
Table \ref{tab:splits} shows the split of train/dev/test for each treebank. Scores reported for the parser are averaged across ten resampling of \textit{CRM14}, \textit{Tafel} and \textit{Trappen} in Phase III.

\subsection{Parsers}
\label{sec:parsers}

Given the robust performance of \textit{Benepar} in constituency parsing across modern languages \cite{kitaev2019Multilingual} and historical languages \cite{arnardottir2020Neural,kulick2022PennHelsinki}, we choose it for our experiments with Middle Dutch. \textit{Benepar} is a span-based neural parser that assigns a score to every possible span of words with a label of a sentence, and uses a modified version of the CKY algorithm \cite{gaddy2018Whats} to combine these span scores and construct the parse tree with the highest overall score. 
PoS tags are assigned using a separate classifier on top of the encoder output, which is jointly optimized with the span classifier (cf. \citealt{kitaev2018Constituency}).
We use the publicly released code of \citet{kitaev2019Multilingual}.\footnote{https://github.com/nikitakit/self-attentive-parser}

Pretrained embeddings have been shown to improve cross-domain parsing performance \cite{yang-etal-2022-challenges, fried-etal-2019-cross, kitaev2019Multilingual}.
Previous studies further indicate that contextualized embeddings trained on data from time periods closer to the target yield better results \cite{kulick2022PennHelsinki,grobol2022BERTrade}.
Accordingly, we use the dbmdz BERT embeddings \cite{devlin2019BERT}, which are trained on historical Dutch.\footnote{https://huggingface.co/dbmdz/bert-base-historic-dutch-cased}
Due to the discrepancy in time and genre of the dbmdz BERT training data (1618-1879, newspapers) and our Middle Dutch data (14th-15th centuries, legal charters and religious texts), we continue to fine-tune the BERT parameters along with parser training, as in the original implementation of \textit{Benepar} \cite{kitaev2019Multilingual}.
\citet{li-etal-2020-semi, li-etal-2022-semi} further show that continued pre-training of BERT with target domain raw text prior to parser training can significantly improve domain adaptation.
However, since it relies on substantial unlabeled data and diverse text genres \cite{gururangan-etal-2020-dont, li-etal-2020-semi, li-etal-2022-semi,grobol2022BERTrade}), we leave it for future work.

As a baseline model, we use the \textit{Bikel} parser \cite{bikel2004Intricacies}, a statistical lexicalized PCFG parser, as it is currently used for constituency parsing of Middle Dutch. It extends the models of \citet{collins1996New, collins1997Three, collins1999HeadDriven} with a more flexible and modular implementation, and enables multilingual parsing. The \textit{Bikel} parser is trained and evaluated using 10-fold cross-validation (Phases I, II and III) on \textit{Etstoel} and 10 resampling runs on the target data (Phase III), following the same data splits as \textit{Benepar}, but without using the dev set, since early stopping is not available. Furthermore, since the \textit{Bikel} parser does not support auxiliary task training or joint training with auxiliary languages, we train it using gold PoS tags only and without any auxiliary language. 

\begin{figure*}[!htp]
  \centering
  \subfigure[{\scriptsize Auxiliary task and languages}]{\includegraphics[scale=0.62]{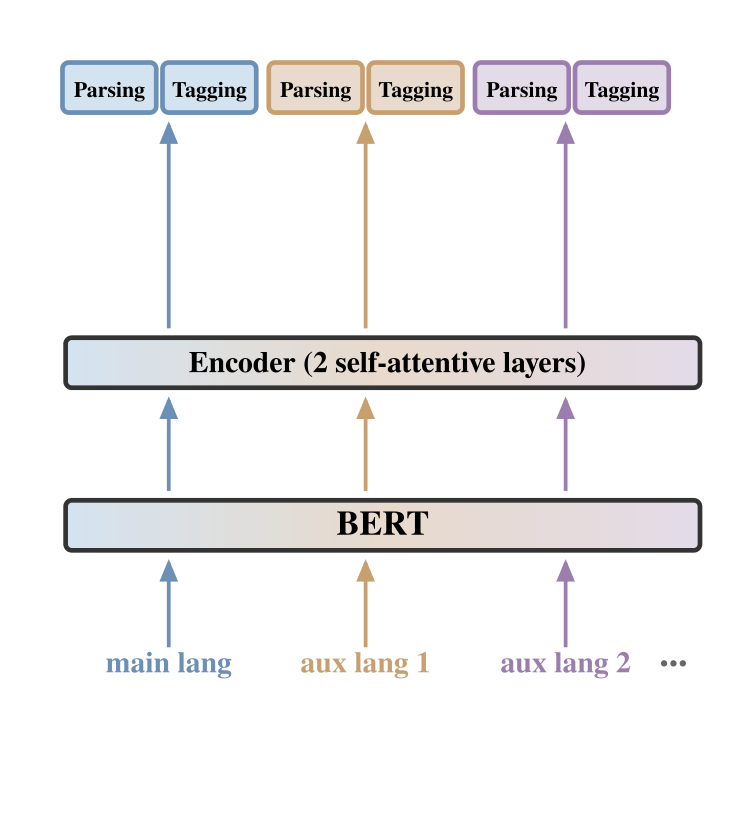}}\quad
  \subfigure[{\scriptsize Fused Representations}]{\includegraphics[scale=0.62]{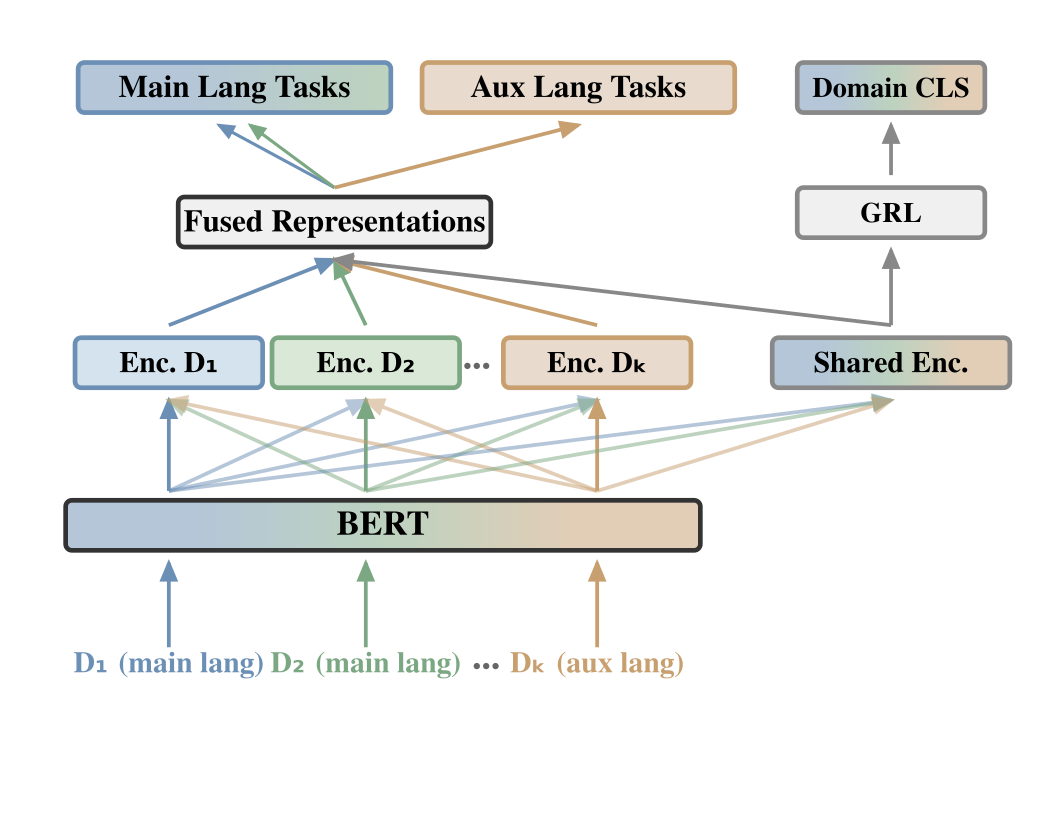}}\quad
  \subfigure[{\scriptsize Multi-source Dynamic Matching}]{\includegraphics[scale=0.62]{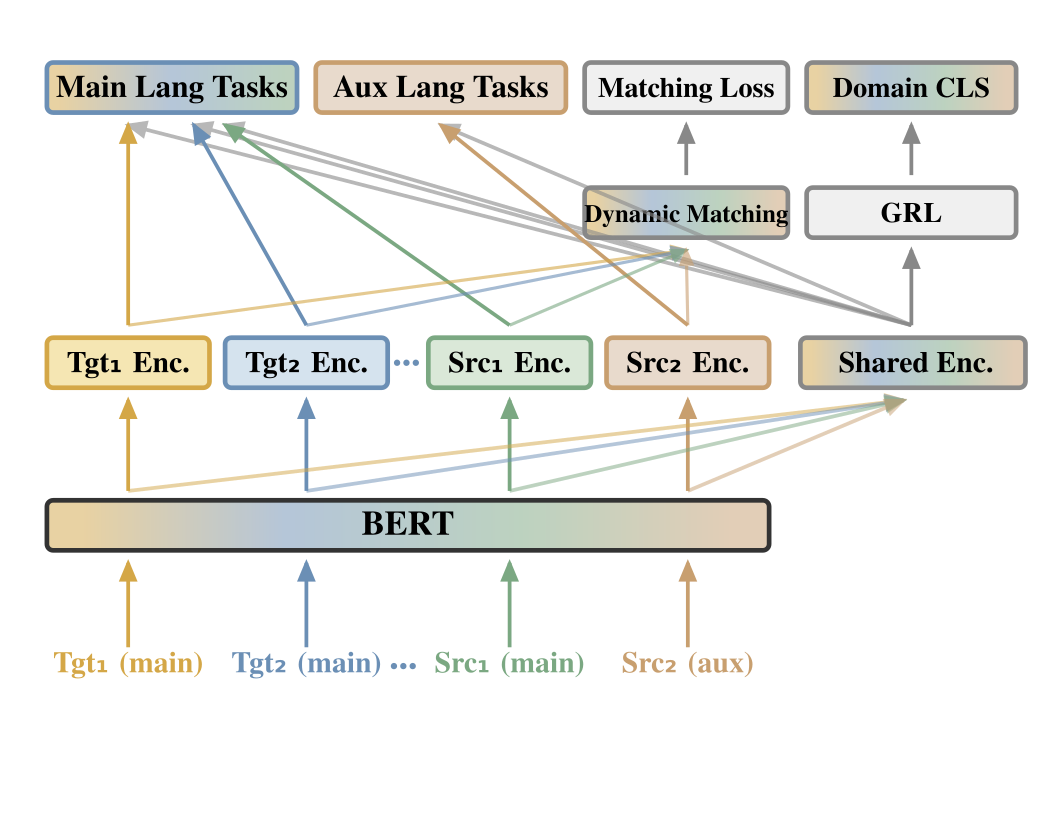}}
  \caption{Model architectures. Enc. stands for encoder and GRL for gradient reversal layer.\label{models}} 
\end{figure*}

\subsection{Phases I: Auxiliary Task}

\citet{coavoux2017Multilingual} have shown that word-level auxiliary tasks improve the performance of their constituency parser based on a bi-LSTM architecture. Therefore, we explore whether training with the PoS-tag prediction as an auxiliary task improves the self-attentive parser for Middle Dutch. In \textit{Benepar} \cite{kitaev2019Multilingual}, two additional randomly initialized self-attention layers are applied to the output of BERT. The parsing module and the optional PoS-tagging module are added on top of these layers as two parallel, independent tasks (cf. Figure \ref{models}a, without auxiliary languages).



\begin{table}[!ht]
\begin{center}
\footnotesize
\begin{tabularx}{\columnwidth}{|p{11em}||p{6em}|r|}
      \hline
      \textbf{Condition} & \textbf{Parsing F1} & $\Delta$ \\
      \hline
      \hline
      Bikel & 71.71 (1.22) & -13.63 \\
      \hline
      Benepar (w/o PoS pred.) & 85.34 (1.08) & 0.00 \\
      \hline
     Benepar (w/ PoS pred.) & 85.48 (1.5) &+0.14\\
      \hline

\end{tabularx}
\caption{\label{tab: results aux tasks} Parsing F1 scores on the \textit{Etstoel} test set, evaluated by evalb, averaged across 10-fold cross-validation, and the difference $\Delta$ with \textit{Benepar} without PoS prediction. Standard deviations are shown in parentheses.}
 \end{center}
\end{table}

\noindent \textbf{Results}~~~Table \ref{tab: results aux tasks} shows the result of 10-fold cross-validation of \textit{Bikel} parser and \textit{Benepar} on \textit{Etstoel}.
\textit{Benepar} significantly outperforms \textit{Bikel} parser regardless of the presence of the auxiliary task, showing that the self-attentive model with pretrained embeddings is more effective in constituency parsing than statistical models, even when trained with less than 2,000 sentences. Regarding the auxiliary task, \textit{Benepar} achieves comparable results when PoS-tag prediction is added or not. This may be due to the self-attentive encoder already implicitly capturing part-of-speech information during parsing. For the remaining experiments, we choose the version with PoS prediction, as this task remains useful for the construction of linguistic treebanks.

\subsection{Phase II: Auxiliary Languages}


\citet{kitaev2019Multilingual} report that incorporating a high-resource language as an auxiliary task can improve parsing performance on lower-resource languages, and that the size of the auxiliary language dataset has a greater impact than morphological relatedness. \citet{sapp2023Parsing} also show that the related high-resource CHLG improve the parsing of the target low-resourced ENHG. 
Given the limited size of parsed Middle Dutch data, we investigate whether existing large-scale constituency treebanks in other languages improve Middle Dutch parsing. 

Unlike \citet{sapp2023Parsing}, who combine source language and target language and use a single parsing module, we assign a separate parsing module to each language and train them jointly, so that Middle Dutch and each auxiliary language maintain distinct tagsets and parameters for parsing, as shown in Figure \ref{models}a, following \citet{kitaev2019Multilingual}.\footnote{We implement this functionality following \citet{kitaev2019Multilingual}, as it is missing from their public Github code.} However, instead of using a multilingual BERT as \citet{kitaev2019Multilingual}, we employ a historical Dutch BERT model (see Section \ref{sec:parsers}), as our goal is to maximize Middle Dutch performance rather than build a universal multilingual parser. The tested auxiliary languages are described in Sectoin \ref{sec:aux lang data}, chosen by data sizes and relatedness to Middle Dutch.
For each auxiliary language, we treat \textit{Etstoel} as the main language (main lang) and perform bilingual joint training. 
For languages with multiple historical stages, we additionally conduct multilingual training in which each auxiliary language has a separate PoS-tagging and parsing module (e.g., MHG = aux lang 1, ENHG = aux lang 2, in Figure \ref{models}a).
The data splits for \textit{Etstoel} and auxiliary languages are detailed in Section \ref{sec:cross validation}, and we apply 10-fold cross-validation as in Phase I.


\begin{table*}[!ht]
\centering
\scriptsize
\setlength{\tabcolsep}{4pt}
\renewcommand{\arraystretch}{1.1}

\begin{tabularx}{\textwidth}{l*{16}{>{\centering\arraybackslash}X}}
\toprule
\textbf{Aux} & ENHG & EModE & MHG & AllHE & CHLG & AllHG & MF & ME & AllHF & OF & HP & Lassy & ModBE & PTB & OE & CTB\\
\# train sents & 11,777 & 28,403 & 6,504 & 267,941 & 6,332 & 18,281 & 39,004 & 79,540 & 110,368 & 71,364 & 55,374 & 31,877 & 54,245 & 45,207 & 105,742  & 64,773\\
\midrule
On Aux &
78.80\newline (0.92) &
84.17\newline (0.72) &
81.31\newline (1.21) &
82.97\newline (1.66) &
83.30\newline (0.73) &
78.75\newline (3.29) &
86.08\newline (0.99) &
83.16\newline (1.52) &
83.61\newline (1.38) &
83.84\newline (0.75) &
83.75\newline (0.58) &
80.82\newline (0.68) &
86.46\newline (1.44) &
\textbf{89.77}\newline (0.61) &
82.73\newline (1.54) &
14.96\newline (1.27) \\

\midrule
\textbf{On \textit{Etstoel}} &
\textbf{86.21}\newline (1.13) &
\textbf{86.21}\newline (1.19) &
86.15\newline (1.05) &
86.14\newline (0.95) &
86.13\newline (1.19) &
86.12\newline (1.12) &
86.08\newline (1.22) &
86.01\newline (1.24) &
86.00\newline (1.06) &
85.97\newline (1.17) &
85.81\newline (1.22) &
85.78\newline (1.28) &
85.77\newline (1.31) &
85.54\newline (1.29) &
85.51\newline (1.30) &
85.50\newline (1.27) \\
$\Delta$ NoAux(85.48)&
\textbf{+0.73} & 
\textbf{+0.73} &
+0.67 &
+0.66 &
+0.65 &
+0.64 &
+0.60 &
+0.53 &
+0.52 &
+0.49 &
+0.33 &
+0.30 &
+0.29 &
+0.06 &
+0.03 &
+0.01 \\
\bottomrule
\end{tabularx}

\caption{\label{tab: aux lang} Parsing F1 scores on auxiliary language test sets and the \textit{Etstoel} test set, ordered by F1 difference $\Delta$ on \textit{Etstoel} when \textit{Benepar} is trained on \textit{Etstoel} train with \textit{vs.} without each auxiliary language (last row). Averages over 10-fold cross-validation; standard deviations in parentheses.}

\end{table*}

\noindent \textbf{Results}~~~Table \ref{tab: aux lang} summarizes the parsing F1 of \textit{Etstoel} without auxilary language, or with different combination of auxiliary languages.
We find that auxiliary languages consistently improve parsing on \textit{Etstoel}, but to varying degrees. 
The top four auxiliary languages are Early New High German (ENHG), Early Modern English (EModE), Middle High German (MHG) and Middle Low German (CHLG). They are all geographically and temporally closest to Middle Dutch, suggesting that morphological proximity contributes greatly to effective transfer.
By contrast, modern languages such as Chinese (CTB), Modern English (PTB and ModBE), and Modern Dutch (Lassy) yield the smallest gains, likely due to differences in the temporal gap from Middle Dutch.
Old English (OE) offers only a weak benefit (+0.03), whereas Old French (OF)'s improvement is much larger (+0.49), which could also be explained by temporal proximity: OE ends around the 11th century, while OF last to the 13th century, closer to \textit{Etstoel}'s composition time (15th c.).
Our findings indicate that temporal and geographical relatedness outweigh corpus size: although ModBE, PTB, and OE are among the largest auxiliary datasets, their impact is minimal, MHG and CHLG, each more than nine times smaller, rank among the most effective.
Overall, our results confirm \citet{kitaev2019Multilingual} that auxiliary languages enhance target language parsing accuracy, but we find that temporal and geographical proximity play a more decisive role than data size.

\subsection{Phase III: Domain adaptation}
After obtaining a good model on single source text, we are interested in finding out ways to improve the parser's cross-domain generalization with data of high heterogeneity and limited quantity.
We first test our best model on \textit{Etstoel} obtained after Phase II directly on the three target domains (CRM, Tafel and Trappen) without any further adaptation (\textbf{0-shot}).
We then fine-tune this model with target domain data in a few-shot learning scheme (\textbf{fine-tune}) as suggested by \citet{chen-etal-2020-low}.
Following \citet{sapp2023Parsing}, we also attempt with models retrained with combined data: Etstoel + one or multiple of the target domains using the same auxiliary language that yield the best model in Phase II (\textbf{combined}).
Additionally, we compare a \textit{Bikel} parser baseline trained with the same combined data (\textbf{Bikel combined}).
Finally, we apply feature separation techniques with adversarial training (adapted from \citealp{sato-etal-2017-adversarial, li-etal-2020-semi, li-etal-2022-semi}) and retrain with combined data and the best auxiliary language (\textbf{DA-fs} and \textbf{DA-msdm}).
The domain adaptation (\textbf{DA}) models are detailed below.

\begin{table*}[!tbp]
\begin{center}
\scriptsize
\begin{tabular}{|c||c|*{1}{p{2.1cm}|}c|*{4}{p{2.1cm}|}}
\hline
\multirow{2}{*}{\textbf{Domain}} & \multicolumn{2}{c|}{\textbf{Bikel}} & \multicolumn{5}{c|}{\textbf{Benepar}} \\
\cline{2-8}
 & \textbf{0-shot} & \textbf{combined} & \textbf{0-shot} & \textbf{fine-tune} & \textbf{combined} & \textbf{DA-fs} & \textbf{DA-msdm} \\
\hline
\textbf{CRM} & 37.32 & 43.34-58.83-62.91 &
    50.69 & \textbf{58.89}-\textbf{71.97}-74.39 & 
    57.84-71.64-74.76 &
    58.36-71.51-\textbf{75.45} & 49.29-71.50-75.31 \\
\hline
\textbf{Tafel} & 39.12 & 45.14-53.25-56.18 &
    65.35 & \textbf{67.02}-71.97-73.64 &
    66.50-\textbf{72.24}-74.16 & 
    66.81-72.11-\textbf{74.73} & 59.27-71.27-74.09 \\
\hline
\textbf{Trappen} & 44.00 & 47.63-55.82-59.10 &
    65.91 & 68.37-\textbf{76.87}-79.09 & 
    68.35-76.48-78.70 & 
    \textbf{68.52}-76.08-\textbf{79.55} & 60.41-75.81-79.06 \\
\hline
\end{tabular}
\caption{\label{tab:resda} Parsing F1 on target domain test sets. Models are trained with 10-100-200 examples from \textit{every} target domain (except for zero-shot; plus \textit{Etstoel} for data-combined and domain adaptation models). Averages over 10-fold cross-validation.}
 \end{center}
\end{table*}

\subsubsection{Adversarial Feature Separation}

We study two methods for low-resource domain adaptation in the adversarial feature separation scheme.
As illustrated in Figure \ref{models}b and \ref{models}c, instead of a single common encoder, a separate encoder is used for each domain plus a shared encoder for all domains.
A domain classifier that receives the output of the shared encoder is trained in an adversarial manner with \textit{gradient reversal} to encourage the shared encoder to learn knowledge not specific to any particular domain \cite{sato-etal-2017-adversarial, ganin2015udabackprop, goodfellow2014gan}.

\paragraph{Class-balanced Batching.}
Our target domain training data are limited in size, especially compared to that of the auxiliary language.
To ensure that the stability of adversarial training, for all the \textbf{DA} models, we use separate class-balanced batching for the domain classifier so that each batch contains an equal amount of input of from each domain.

\paragraph{Orthogonality Constraints.}
Following \citet{bousmalis2016domainseparationnetworks, li-etal-2020-semi, li-etal-2022-semi}, we apply \textit{orthogonality constraints} to encourage each private encoder to learn different representations than the shared one.
The loss of orthogonality constraints is defined as follows:
\begin{equation}
\setlength\abovedisplayskip{3pt}
\setlength\belowdisplayskip{3pt}
\mathcal{L}_{\text{ort}}
=
\frac{1}{N}
\sum_{d \in \mathcal{D}}
\left\|
\left( \mathbf{H}_c \right)^\top
\mathbf{H}_p^{(d)}
\right\|_F^2
\end{equation}
where $\|\cdot\|^2_F$ is the squared Frobenius norm, $\mathcal{D}$ is the set of $N$ domains, $\mathbf{H}_c$ is the shared representations, and $\mathbf{H}_p^{(d)}$ is the domain $d$ private representations.
The orthogonality constraints are implemented for all the \textbf{DA} models.

\paragraph{Fused Representations.}
In the shared-private scheme, the parsing module leverages representations from both the shared and private encoders.
\citet{sato-etal-2017-adversarial} use gates to mix representations, but observe harm to performance when target domain data are scarce.
This may be due to underfit of the target private encoder and of the target-specific gate parameters.
\citet{li-etal-2020-semi} propose to alleviate this problem with \textit{fused target-domain word embeddings}, which uses mixed private representations (from both target and source domain encoders) when the input comes from the low-resource target domain, and keeps the raw private representations if the input comes from the high-resource source domain.
The rational behind this is to leverage transferable knowledge from a better optimized encoder of a similar domain trained with sufficient data.
However, the ratio of this mixture is treated as a hyperparameter, whose choice can be crucial to adaptation performance.

In light of this, we propose to fuse the outputs of different encoders using a set of learnable coefficients (cf. Figure \ref{models}b):
\begin{equation}
\setlength\abovedisplayskip{3pt}
\setlength\belowdisplayskip{3pt}
\mathbf{H}^{(d')} 
= \alpha_c^{(d')} \, \mathbf{H}_c 
 + \sum_{d \in \mathcal{D}} \alpha_p^{(d',d)} \, \mathbf{H}_p^{(d)}
\end{equation}
where $\mathbf{H}^{(d')}$ denotes the fused representations for domain $d' \in \mathcal{D}$ fed into the task networks, $\alpha_c^{(d')}$ and $\alpha_p^{(d',d)}$ are domain-specific coefficients for the shared and private representations respectively, and $\alpha_c^{(d')} + \sum_{d \in \mathcal{D}} \alpha_p^{(d',d)} = 1$.
The initial values of these coefficients are set proportional to data sizes. 
We implement this feature in the \textbf{DA-fs} model. 

\paragraph{Multi-source Dynamic Matching.}
\citet{li-etal-2022-semi} tackle the low-resource private encoder underfit problem from a similar angle.
They use a \textit{dynamic matching network} to encourage the target encoder to learn from useful source features. 
We make simple extension to this network in order to adapt to our multi-domain and multi-source setting.\footnote{We refer the reader to \citet{li-etal-2022-semi} for detailed descriptions of the domain matching network design.}
The modified matching loss is defined as follows:
\begin{small}
\begin{equation}
\setlength\abovedisplayskip{3pt}
\setlength\belowdisplayskip{3pt}
\begin{aligned}
\hspace{-1em}
\mathcal{L}_{\text{mat}}
&= \frac{1}{|T||S|}\sum_{i \in T} \sum_{j \in S}
    \, \mathcal{L}_{\text{mat}}^{(i,j)}, \\
\mathcal{L}_{\text{mat}}^{(i,j)}
&= \frac{1}{K D}
   \sum_{n,m}
   W_{i,j}^{n,m} \,
   \sum_{d=1}^{D}
   Q_{i,j,d}^{n,m}
   \bigl( f_\theta(t_{j}^m) - s_{i}^n \bigr)^2_d
\end{aligned}
\end{equation}
\end{small}%
where \(S\) and \(T\) denote the sets of source (teacher) and target (student) domain encoders, respectively.  
\(i \in T\) and \(j \in S\) index a target and a source encoder.  
\(W_{i,j}^{n,m}\) and \(Q_{i,j,d}^{n,m}\) are layer- and element-level matching weights.  
\(f_\theta(\cdot)\) is a linear transformation. 
\(s_{i}^n\) and \(t_{j}^m\) denote outputs of the \(n\)-th and \(m\)-th layers of encoders \(E_i\) and \(E_j\).  
\(D\) is the encoders' output dimension, and \(K\) the total number of layer pairs.
The multi-source domain matching loss is implemented by the \textbf{DA-msdm} model. 


\subsubsection{Results}

Table \ref{tab:resda} summarizes the parsing F1 of different models on the three target domain test sets (CRM, Tafel and Trappen), where the Benepar-fine-tune model is fine-tuned on 10, 100 or 200 examples from \textit{every} target domain (thus 30, 300 or 600 examples in total), and plus \textit{Etstoel} for data-combined and domain adaptation models.
We find that \textit{Benepar} strongly outperforms \textit{Bikel} parser even in zero-shot conditions.
The \textit{Benepar} zero-shot model even outperforms data-combined \textit{Bikel} with 200 examples each when tested on \textit{Tafel} and \textit{Trappen}, or that with 10 examples each when tested on \textit{CRM}.
Fine-tuning and data-combined retraining with sentences from the target domain yield comparable results, and the parser begins to show improvement with as few as 10 examples each and surpasses 70 F1 after 100 examples each.

We further examine feature separation techniques for domain adaptation and find that, with around 200 examples each per domain, cross-domain performance exceeds 74.7 F1 across all three new domains, while smaller datasets yield little or no improvement.
Having only 10 examples from every target domain is detrimental to paring performance in the case of the model with multi-source dynamic matching (DA-msdm).
This is likely because of the lack of training examples in the target domains causing training instabilities of the matching network.

\section{Conclusion}

We have adapted a transformer-based constituency parser to Middle Dutch and have shown that jointly training with temporally, geographically and typologically closer auxiliary languages yields larger improvements.
For new domains, fine-tuning offers a faster alternative to data-combined retraining with similar effectiveness: small gains appear with about 10 examples, and clear improvements begin around 100.
Feature separation also helps cross-domain performance but only when at least 200 sentences are available.

Future work includes exploiting unlabeled Middle Dutch data through continued pretraining (cf. \citealp{li-etal-2020-semi, li-etal-2022-semi}; see also \citealp{gururangan-etal-2020-dont, grobol2022BERTrade}) or semi-supervised learning (cf. \citealp{rotman-reichart-2019-deep}) with auxiliary tasks, deepening exploration of domain adaptation with ongoing development of parsed and unparsed texts, and exploring multilingual objectives for historical languages.

\newpage

\section{Bibliographical References}
\bibliographystyle{lrec2026-natbib}
\bibliography{lrec2026-example}

\end{document}